\documentclass[11pt]{article}

\usepackage[preprint]{acl}
\usepackage{amsmath}
\usepackage{ascmac}
\usepackage{float}
\usepackage{times}
\usepackage{booktabs}
\usepackage{latexsym}
\usepackage{subcaption}
\newcommand{\ctext}[1]{\raise0.2ex\hbox{\textcircled{\scriptsize{#1}}}}

\usepackage[T1]{fontenc}

\usepackage[utf8]{inputenc}

\usepackage{microtype}

\usepackage{inconsolata}

\usepackage{graphicx}
\usepackage{enumitem}
\usepackage {arydshln}
\title{A Japanese Language Model and Three New Evaluation Benchmarks for Pharmaceutical NLP}

\author{
 \textbf{Shinnosuke Ono\textsuperscript{1,2,*}},
 \textbf{Issey Sukeda\textsuperscript{1,2,*}},
 \textbf{Takuro Fujii\textsuperscript{1}},
 \textbf{Kosei Buma\textsuperscript{1,3}},
 \textbf{Shunsuke Sasaki\textsuperscript{1,2}},
\\
\\
 \textsuperscript{1}EQUES Inc.,
 \textsuperscript{2}The University of Tokyo,
 \textsuperscript{3}University of Tsukuba,
\\
}

\begin{document}
\maketitle
\def\thefootnote{*}\footnotetext{These authors contributed equally to this work}\def\thefootnote{\arabic{footnote}}

\begin{abstract}
We present a Japanese domain-specific language model for the pharmaceutical field, developed through continual pretraining on 2 billion Japanese pharmaceutical tokens and 8 billion English biomedical tokens. To enable rigorous evaluation, we introduce three new benchmarks: \textbf{YakugakuQA}, based on national pharmacist licensing exams; \textbf{NayoseQA}, which tests cross-lingual synonym and terminology normalization; and \textbf{SogoCheck}, a novel task designed to assess consistency reasoning between paired statements.
We evaluate our model against both open-source medical LLMs and commercial models, including GPT-4o. Results show that our domain-specific model outperforms existing open models and achieves competitive performance with commercial ones, particularly on terminology-heavy and knowledge-based tasks. Interestingly, even GPT-4o performs poorly on SogoCheck, suggesting that cross-sentence consistency reasoning remains an open challenge.
Our benchmark suite offers a broader diagnostic lens for pharmaceutical NLP, covering factual recall, lexical variation, and logical consistency. This work demonstrates the feasibility of building practical, secure, and cost-effective language models for Japanese domain-specific applications, and provides reusable evaluation resources for future research in pharmaceutical and healthcare NLP.
Our model, codes, and datasets will be released upon acceptance.
\end{abstract}

\section{Introduction} \label{introduction}
\begin{figure}[t]
    \centering
    \includegraphics[width=\linewidth]{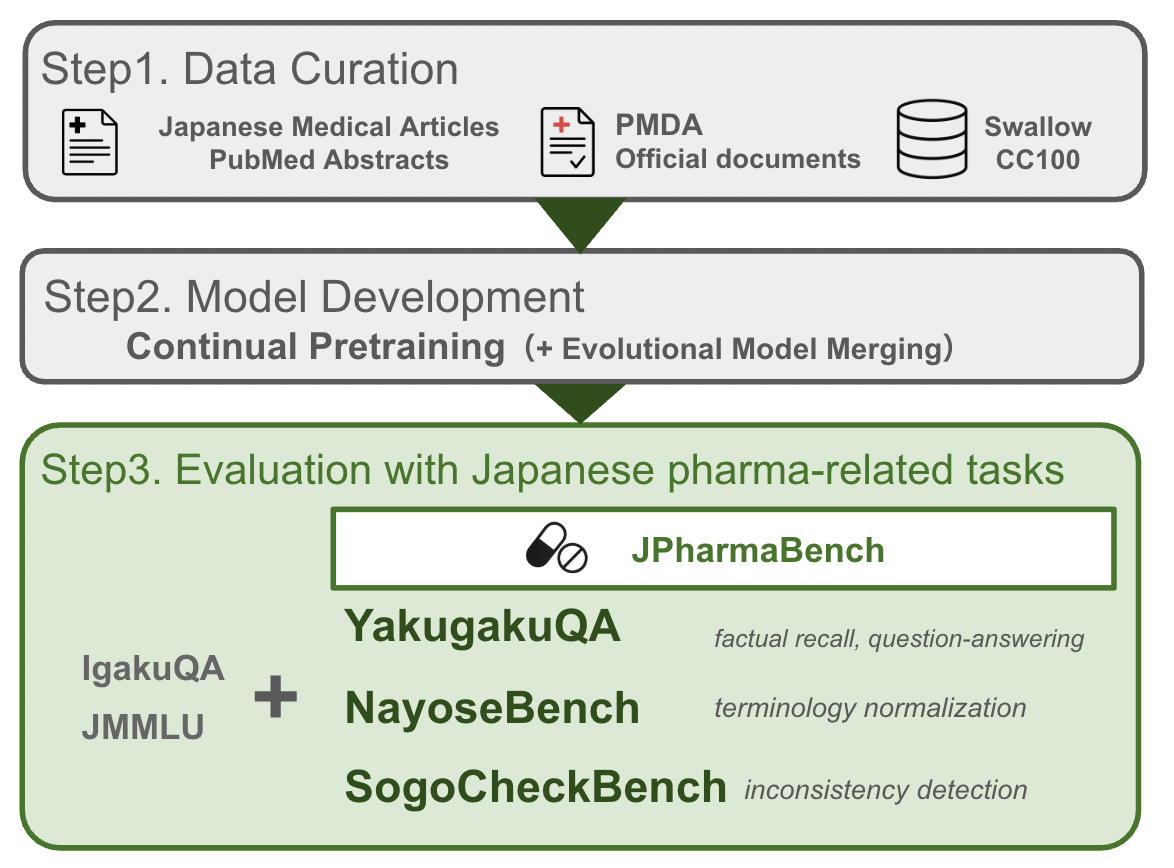}
    \caption{\textbf{\textsc{JPharmatron} and \textsc{JPharmaBench}.} The pipeline for data curation, continued pretraining, and evaluation of \textsc{JPharmatron}.}
    \label{fig:overview}
\end{figure}

\begin{figure}[t]
    \centering
    \includegraphics[width=\linewidth]{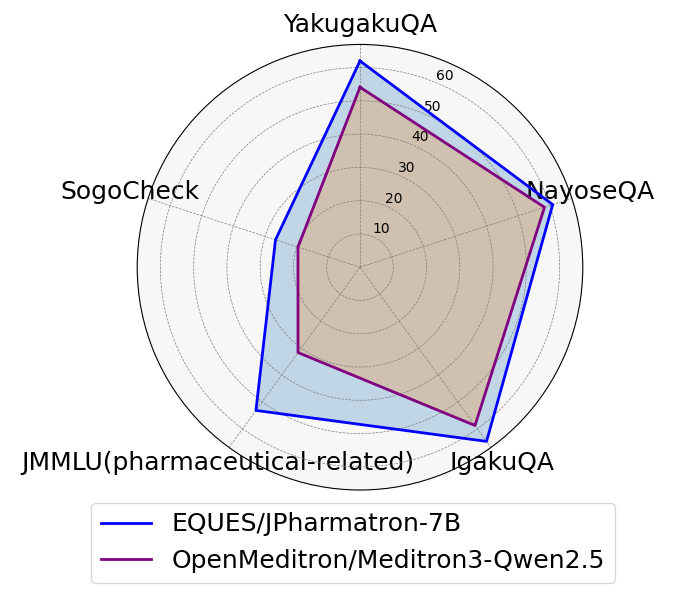}
    \caption{\textbf{Performance Comparison with Meditron.}  \textsc{JPharmatron} consistently achieves higher scores than Meditron across \textsc{JPharmaBench}, IgakuQA, and JMMLU.}
    \label{fig:chart}
\end{figure}

Large Language Models (LLMs) have achieved remarkable performance across a wide range of natural language processing (NLP) tasks. However, their effectiveness remains limited in domain-specific settings such as manufacturing, finance, and medicine~\cite{islam2023financebench,hager2024evaluation,zhang2024potential}, where deep contextual understanding and precise terminology handling are required. In these domains, general-purpose LLMs often fall short due to inadequate domain knowledge and difficulty handling complex or specialized queries. Moreover, while domain-specific fine-tuning can enhance surface-level performance, it has been shown that this does not necessarily lead to genuine knowledge acquisition~\cite{zhou2023lima}.

The pharmaceutical domain is no exception. In particular, the Japanese pharmaceutical industry faces significant administrative overhead in tasks such as document preparation, verification, and regulatory compliance—often governed by standards such as GMP~\cite{gmp} and ICH guidelines\footnote{\url{https://www.ich.org/page/ich-guidelines}}. Despite these challenges, little work has been done to develop LLMs tailored for pharmaceutical operations, especially in Japanese.

In this work, we present \textsc{JPharmatron}, a Japanese language LLM series specialized for pharmaceutical operations. To build \textsc{JPharmatron}, we perform continual pretraining of the Qwen2.5~\cite{yang2024qwen2} model using a curated corpus consisting of Japanese pharmaceutical journals, web resources, and synthetic data (Appendix~\ref{appendix:model_training}). Unlike prior work focusing on drug discovery~\cite{chaves2024txllmlargelanguagemodel, tsuruta2024a}, our model targets real-world operational tasks, such as document standardization and terminology normalization.

To evaluate pharmaceutical reasoning and generation capabilities, we introduce three novel benchmarks:

(1) YakugakuQA (\S\ref{yakugakuqa}): a multiple-choice QA dataset based on the Japanese National Pharmacist Examination;

(2) NayoseQA (\S\ref{nayosebench}): a paraphrasing benchmark for standardizing drug names and active substances;

(3) SogoCheck (\S\ref{sogobench}): a document consistency-check task reflecting real administrative workflows.

These benchmarks, collectively referred to as \textsc{JPharmaBench}, are designed to reflect practical scenarios encountered in pharmaceutical companies, particularly in regulatory and clerical operations. To the best of our knowledge, this is the first benchmark suite for evaluating LLMs in Japanese pharmaceutical applications.

We evaluate \textsc{JPharmatron} using in-context learning across \textsc{JPharmaBench} and two existing benchmarks additionally. Without task-specific fine-tuning, our model outperforms competitive LLMs including Meditron (\S\ref{meditron}), showing gains of 7.9\% on YakugakuQA (Ours) and 5.9\% on IgakuQA~\cite{jpn-med-exam_gpt4}. These results suggest that domain-adaptive continual pretraining can significantly enhance LLM performance in specialized pharmaceutical and medical settings.

Our contributions are threefold:
\begin{itemize}[itemsep=0pt, topsep=3pt]
    \item We introduce the first LLMs and evaluation benchmarks specifically designed for Japanese pharmaceutical NLP.
    \item We develop tasks aligned with real-world workflows, ensuring practical relevance in pharmaceutical operations.
    \item We provide a complete methodology --- from data collection to evaluation --- that serves as a replicable and secure framework for domain-specific LLM development in regulated industries.
\end{itemize}

\section{Related works}

\subsection{Domain-specific LLMs and benchmarks in healthcare}

With the emergence of GPTs~\cite{radford2018improving,brown2020language}, domain-specific adaptations for healthcare have rapidly gained attention. Several English-centric LLMs have been developed to infuse medical knowledge into general-purpose models. For instance, Med-PaLM 2~\cite{singhal2023towards}, a specialized version of PaLM 2~\cite{anil2023palm2technicalreport}, is fine-tuned on curated medical datasets and achieves performance comparable to medical professionals on exams.

Benchmarking has evolved in parallel. MultiMedQA~\cite{singhal2023large} combines datasets to evaluate both factual knowledge and clinical reasoning. Other benchmarks, such as MedQA~\cite{jin2020disease} and the medical subset of MMLU~\cite{hendryckstest2021}, are commonly used to assess instruction-following and medical understanding.

In the Japanese context, GPT-style healthcare LLMs are still emerging. Recent projects~\cite{sukeda2023jmedlora, sukeda202470b, sukeda2024development} have focused on adapting LLMs for Japanese medical question answering. The standard benchmarks are also being developed~\cite{sukeda2024development2, jiang2024jmedbenchbenchmarkevaluatingjapanese}, exemplified by IgakuQA~\cite{jpn-med-exam_gpt4}, based on the Japanese national medical licensing exam.

These developments in both English and Japanese highlight a global trend toward aligning LLMs with clinical expertise across languages and contexts. While significant progress has been made in the medical field, efforts in the pharmaceutical domain remain limited, and the few existing models~\cite{chen2024pharmagpt,chaves2024txllmlargelanguagemodel} are not publicly available.

\subsection{Meditron} \label{meditron}

Among existing domain-specific medical LLMs, Meditron~\cite{chen2023meditron} is particularly relevant to our work. Meditron is a family of open-source LLMs of 7B and 70B, built upon LLama2~\cite{touvron2023llama}, and adapted with medical continual pretraining and supervised fine-tuning using curated English medical corpus. It demonstrates strong performance in MedQA~\cite{jin2020disease}, making it a prominent example of an open medical LLM. The work is further extended by Open Meditron Initiative\footnote{\url{https://huggingface.co/OpenMeditron}}.

In contrast, our work focuses on the Japanese language and the pharmaceutical domain, both of which remain underexplored. With strong performance on YakugakuQA, our model serves as a Japanese-pharmaceutical counterpart to Meditron. This parallel extends to benchmarks as well: Meditron is evaluated on MedQA~\cite{jin2020disease}, while our model is evaluated on YakugakuQA (ours) and IgakuQA~\cite{jpn-med-exam_gpt4}, which are all based on national licensing exams in their respective languages and domains.

\section{Benchmark construction}

Pharmaceutical domain has not received as much attention for LLM applications, resulting in a limited number of evaluation benchmarks, especially in Japanese.
When the focus is solely on therapeutics data, a comprehensive benchmark for therapeutics machine learning called the \textit{Therapeutic Data Commons}~\cite{Huang2022artificial} can be applied to LLM evaluations~\cite{chaves2024txllmlargelanguagemodel}. However, the performance of LLMs in the broader pharmaceutical domain has only been evaluated on the North American Pharmacist Licensure Examination (NAPLEX)~\cite{ehlert2024large, chen2024pharmagpt}, with no evaluations conducted in Japanese.
Although MMLU~\cite{hendryckstest2021} and JMMLU~\cite{yin2024should} cover related healthcare domains, neither includes pharmaceutics as a distinct category.

\subsection{Overview of \textsc{JPharmaBench}}
To evaluate language models in the Japanese pharmaceutical domain, we constructed three novel benchmarks, each reflecting a different type of reasoning or knowledge required in real-world pharmaceutical practice: factual recall, terminology normalization, and inconsistency detection (Table~\ref{tab:benchmark_comparison}).
All benchmarks are based on publicly available data and are structured as question-answering tasks, making them compatible with various LLMs. 

\begin{table*}[t]
    \centering
    \resizebox{\linewidth}{!}{
    \begin{tabular}{l l l l r l}
    \toprule
    \textbf{Benchmark} & \textbf{Format} & \textbf{Main Skill} & \textbf{Source} & \textbf{\#Examples} & \textbf{Language(s)} \\ \midrule
    YakugakuQA & 4-to-6-choice QA & Factual recall & Licensing exams & 3,021 & Japanese \\
    NayoseQA   & 5-choice QA & Terminology normalization & KEGG DRUG Database & 34,769 & Japanese / English \\
    SogoCheck  & Sentence pair & Inconsistency detection & Japanese Pharmacopoeia & 200 & Japanese \\
    \bottomrule
    \end{tabular}}
    \caption{\textbf{An overview of \textsc{JPharmaBench}, the three pharmaceutical benchmarks for evaluation.} Each task is designed to assess different capabilities of LLMs in domain-specific settings.}
    \label{tab:benchmark_comparison}
    \vspace{-2mm}
\end{table*}

\subsection{YakugakuQA: National Licensing Exam} \label{yakugakuqa}

YakugakuQA is a question-answering dataset based on the Japanese national pharmacist licensing examinations (NPLE) administered by the Ministry of Health, Labour and Welfare. As illustrated in Figure~\ref{fig:example-question}, each question requires selecting one or two correct answers from five or six choices. As summarized in Table~\ref{tab:overview}, YakugakuQA serves as a pharmaceutical counterpart to IgakuQA.

\begin{figure}[!t]
    \centering
    \begin{screen}
        \textit{Which of the following is not an ideal property of a dilute solution? Choose one.}\\
        1. \textit{Vapor pressure lowering}\\
        2. \textit{Freezing point depression}\\
        3. \textit{Boiling point elevation}\\
        4. \textit{Surface tension reduction}\\
        5. \textit{Osmotic pressure}
    \end{screen}
    \caption{\textbf{An example question from the Japanese National Pharmacist Licensing Examination.} The model is required to output ``4'' in this case. The question is originally in Japanese, but translated into English by ChatGPT for readability.}
    \label{fig:example-question}
\end{figure}
\begin{table}[!t]
    \centering
    \resizebox{0.9\linewidth}{!}{
    \begin{tabular}{c|cc}
        \toprule
         & \textbf{English} & \textbf{Japanese}\\
         \midrule
         Medicine& MedQA& IgakuQA \\
         &\cite{jin2020disease}  & \cite{jpn-med-exam_gpt4}\\ 
         \midrule
         Pharmacy& NAPLEX & {\bf YakugakuQA}\\
         & (not structured) &{\bf (Ours)}\\
         \bottomrule
    \end{tabular}
    }
    \caption{\textbf{National licensing exams.} These are typically used as benchmarks when evaluating domain-specific LLMs in medical-related fields.}
    \label{tab:overview}
    \vspace{-2mm}
\end{table}

We have collected the exam data from the past 13 years, from 2012 to 2024.
All questions, answers, and commentaries have been obtained from the website \textit{yakugaku lab}\footnote{\url{https://yakugakulab.info/}} and manually processed. 
The category varies among pharmacy and eight other related areas: pharmacy, pharmacology, chemistry, pathology, hygiene, physics, practice, law, and biology.

Some questions in the NPLE require responses based on a provided image --- for example, identifying a chemical reaction depicted in the image.
However, such image-based questions are excluded from our experiments, as our study focuses on LLMs designed for text input.
The number of questions by year and category used in our experiments is shown in Table~\ref{tab:number-of-yakugakuqa}.

\subsection{NayoseQA: Synonym and Terminology Normalization in the Pharmaceutical Domain} \label{nayosebench}
NayoseQA is our original benchmark designed to evaluate LLMs' ability to handle lexical variation and term normalization in pharmaceutical texts written in Japanese. The task focuses on resolving different surface forms of the same underlying drug or chemical entity, including:

\begin{itemize}[itemsep=0pt, topsep=3pt]
    \item Japanese name $\leftrightarrow$ English name \\
    \item brand name $\leftrightarrow$ generic name \\(e.g., Ganaton $\leftrightarrow$ Itopride hydrochloride)
    \item chemical name $\leftrightarrow$ common name \\(e.g., Prostaglandin E2 $\leftrightarrow$ PGE2)
\end{itemize}

\noindent This type of normalization is commonly referred to as ``nayose'' in Japanese, a term used in information systems to describe the process of identifying and consolidating records that refer to the same real-world entity. In our context, it involves linguistic and domain-specific reasoning to recognize synonymous terms for pharmaceutical compounds.
In real-world pharmaceutical documents and practice in Japan, such variations are common due to regulatory terminology, manufacturer-specific branding, and historical naming conventions. Accurately interpreting and normalizing these variations is essential for drug interaction checks, medical record standardization, and multilingual information retrieval.

\subsection{SogoCheck: Inconsistency Detection in Paired Pharmaceutical Statements} \label{sogobench}

SogoCheck is a novel benchmark we introduce to evaluate an LLM's ability to detect logical or factual inconsistencies (referred to as "sogo" in Japanese) between two pieces of text in the pharmaceutical domain. Unlike factual question-answering benchmarks, which assess whether a synthetic text contains any factual errors~\cite{zhao2023felm}, SogoCheck focuses on cross-text consistency. The task is inspired by a common practice in pharmaceutical quality assurance in Japan, where experts conduct consistency reviews to cross-validate information across documents such as package inserts, internal quality assurance logs, and regulatory submissions.

In this task, the model is presented with a pair of short Japanese texts, typically drawn from regulatory documents, drug descriptions, or quality assurance manuals. The model is asked to determine whether the two statements are consistent either explicit or implicit. Some examples are clear-cut (e.g., numerical mistakes, see Figure~\ref{fig:example-question-sogo}), while others require pharmacological reasoning or recognition of subtle semantic contradictions.

\begin{figure}[!t]
    \centering
    \begin{screen}
    {\small
        \textbf{Text A:} \textit{Storage method: sealed container. Temperature below 25${}^\circ$C. Humidity below 60\%.}\\
        \textbf{Text B:} \textit{Storage method: sealed container. Temperature below 26${}^\circ$C. Humidity below 61\%.}\\
        \textbf{Label:} Change in temperature and humidity }
    \end{screen}
    \caption{\textbf{A simplest example from SogoCheck.} The numbers are inconsistent across two inputs. Originally in Japanese, but translated for readability.}
    \label{fig:example-question-sogo}
\end{figure}


The final dataset includes 200 examples, synthesized with an LLM to balance clarity and realism. This benchmark is particularly valuable because inconsistency detection is crucial in practical workflows such as regulatory review, where conflicting information can lead to severe medical or legal consequences.

\section{Model \& Training}
We developed a domain-specific language model, \textsc{JPharmatron}, through continual pretraining with three different data scales, based on Qwen2.5-7B~\cite{yang2024qwen2}, a multilingual open-source language model that also supports Japanese input, and evolutionary merging. This base model was chosen for its strong general performance, multilingual capacity, and availability under a commercially permissible license.

To inject domain-specific knowledge while preserving general language capabilities, we adopted continual pretraining rather than training from scratch. We prepared three variations of the training corpus:

\paragraph{2B tokens:} Approximately 2B Japanese tokens sourced from pharmaceutical-related documents such as journal papers and drug package inserts;

\paragraph{10B tokens:} The above 2B Japanese tokens combined with an additional 8B English tokens from PubMed Abstracts;

\paragraph{9B tokens:} Based on the 10B-token corpus, further augmented with 1.2B tokens from the CC100 multilingual dataset. After removing duplicates, the number of tokens was finally 9B tokens (see Appendix~\ref{appendix:model_training} for details).

\noindent Training was conducted using standard autoregressive language modeling objectives with the original tokenizer of Qwen2.5. 
Table~\ref{tab:training} provides an overview of the training configuration and data composition. In addition, model merging was performed to attach instruction-following ability to the model.
Further details on data collection, cleaning, and preprocessing pipelines are defered to Appendix~\ref{appendix:model_training}.

We emphasize that our goal was not to outperform proprietary LLMs like GPT-4o, but to develop a practically deployable model as a first baseline that balances accuracy, efficiency, and privacy for real-world use in Japanese pharmaceutical contexts. This lightweight domain adaptation strategy enables enterprises to build specialized models without large-scale resources (\S\ref{deployable}).

\begin{table}
    \centering
    \resizebox{0.9\linewidth}{!}{
    \begin{tabular}{cc} \toprule
        \multicolumn{2}{c}{\textbf{Training Settings} } \\\midrule
         Method& Continual pretraining\\ 
         Base model& Qwen2.5-7B\\ 
         Japanese data & 2B tokens (pharma-related)\\
         English data & 8B tokens (mainly PubMed Abstracts)\\
         Tokenizer & Qwen2.5 tokenizer\\
         Steps & 67171\\
         Batch size & 16\\
         Optimizer & hybridadam\\
         Learning rate & $1.0\times 10^{-5}$\\
         GPU& 8 $\times$ NVIDIA H100\\ 
         Framework& Pai-Megatron-Patch\\ 
         GPU hours& 444\\\bottomrule
    \end{tabular}
    }
    \caption{\textbf{Details of model training settings.}}
    \label{tab:training}
\end{table}

\section{Evaluation}

\subsection{Experimental Setups}

We evaluated our domain-specific model against three types of baseline models: (1) a general-purpose Japanese LLM (Swallow series or equivalent), (2) a medical LLM (Meditron)\footnote{We use Meditron3-Qwen2.5-7B from OpenMeditron for comparison, as the older version~\cite{chen2023meditron} lacks sufficient Japanese support and our model is also based on Qwen2.5-7B, ensuring a fair evaluation.}, and (3) GPT-4o via the OpenAI API. Evaluation was conducted across three newly proposed benchmarks --- YakugakuQA, NayoseQA, and SogoCheck --- as well as two existing Japanese medical benchmarks: IgakuQA and a pharmaceutical subset of JMMLU. This setup enables direct comparison with prior work.

To ensure fairness, all models were prompted with consistent formatting (details provided in Appendix~\ref{appendix:benchmarks}). For multiple-choice questions, models were instructed to select one or more answer options as appropriate, where the accuracy was measured based on exact match.

\subsection{Quantitative results}

Table~\ref{tab:performance} shows the accuracy of each model on each  benchmark.
While GPT-4o achieved the highest accuracy overall, as expected from a frontier commercial LLM, our domain-specific model consistently outperformed both Meditron and the general-purpose Japanese model across all tasks. This highlights the effectiveness of domain-specific continual pretraining in Japanese, and establishes our model as the strongest open alternative for pharmaceutical NLP tasks in the Japanese language.

Breaking down by benchmark, on YakugakuQA, our model achieved an accuracy of 62.0\%, outperforming Meditron3-Qwen2.5-7B by 7.9 points. This result suggests that factual pharmaceutical knowledge can be effectively captured through continual pretraining, even without training from scratch. 
In addition, it suggests that medical domain specialization alone may be insufficient for handling pharmaceutical tasks effectively.
The accuracy results by categories are listed in Table~\ref{tab:yakugakuqa-by-category}, along with additional larger models for references: Llama-3.1-Swallow-70B~\cite{Fujii:COLM2024}, Qwen2.5-72B-Instruct~\cite{yang2024qwen2}, and o1-preview via OpenAI API. 

In NayoseQA, which tests synonym normalization and cross-lingual terminology mapping, the performance gap between our domain-specific model and the general-purpose model (Llama3.1-Swallow) was surprisingly small. This suggests that the task primarily requires lexical and semantic matching capabilities rather than deep domain-specific pharmaceutical knowledge.
While domain adaptation improved performance modestly, it appears that general LLMs with strong multilingual and synonym handling capabilities can already perform well on such terminology normalization tasks. This indicates that future pharmaceutical LLM development efforts may benefit more from enhancing complex reasoning and factual recall abilities rather than focusing solely on terminology alignment.

Finally, SogoCheck proved to be challenging for all models. While one of our models outperformed Meditron by 7.1 points, the absolute accuracy remained low. Notably, even GPT-4o achieved only 39.1\% accuracy, suggesting that subtle consistency detection in specialized domains remains an open research challenge.
Interestingly, many SogoCheck examples were intentionally designed to be solvable by simple textual comparison --- identifying surface-level differences without requiring deep reasoning (see Figure~\ref{fig:example-question-sogo}). Despite this, LLMs often failed to detect such inconsistencies, indicating that current models still struggle with fine-grained semantic alignment even when superficial textual clues are available. This gap between human intuition and model behavior highlights a critical limitation in today's LLM architectures.

\begin{table*}
    \centering
    \resizebox{0.9\linewidth}{!}{
    \begin{tabular}{llccccc} \toprule
       &\textbf{Model}  & \textbf{YakugakuQA}  & \textbf{NayoseQA} & \textbf{SogoCheck} & \textbf{IgakuQA} & \textbf{JMMLU}\\ \midrule
       (1)&TinySwallow-1.5B-Instruct & 37.2 &35.3& 3.1 &39.0&32.1\\
       &sarashina2.2-3b-instruct &46.2&45.6& 0.66 &41.6&37.8\\
       &Llama-3-Swallow-8B-Instruct-v0.1 &42.6&29.8&-&41.5&20.6\\
       &Llama-3.1-Swallow-8B-Instruct-v0.3&48.2&57.6&-&45.2&44.0\\ \midrule
       (2)&Meditron3-Qwen2.5-7B  & 54.1   & 58.3 & 19.6 & 58.8 &31.7\\ \midrule
       (3)&GPT-4o  & \bf{83.6} &  \bf{86.0} & \bf{39.1} & \bf{86.6} & \bf{79.1}\\ \midrule
       Ours& \textsc{JPharmatron}-7B /2B tokens&60.7&58.3& 12.5 &62.3& \bf{55.0}\\
       & \textsc{JPharmatron}-7B /10B tokens& 54.8& \bf{62.6}& 22.0&60.1& 48.7\\ 
       & \textsc{JPharmatron}-7B /9B tokens&\bf{62.0} &60.9&\bf{26.7}&\bf{64.7}&53.2\\ \bottomrule
    \end{tabular}}
    \caption{\textbf{Performance of our LLMs in five pharmaceutical-related benchmarks}, compared to (1) a general-purpose Japanese LLM (Swallow series, or equivalent), (2) a medical LLM (Meditron), and (3) GPT-4o. Each value shows the accuracy (\%). ``-'' denotes the lack of instruction-following capability to solve each task. The top two models for each task are highlighted in bold.}
    \label{tab:performance}
\end{table*}

\begin{table*}[!ht]
    \centering
    \resizebox{\linewidth}{!}{
    \begin{tabular}{lccccccccc|c}\toprule
          \textbf{Model}& \textbf{Biology} & \textbf{Chemistry} & \textbf{Hygiene} & \textbf{Law} & \textbf{Pathology} & \textbf{Pharmacology} & \textbf{Pharmacy} & \textbf{Physics} & \textbf{Practice} & \textbf{Overall}\\ \midrule
          TinySwallow-1.5B-Instruct &41.1&21.9&34.4&\textbf{\textbf{46.5}}&\textbf{44.3}&27.8&36.9&32.4&38.0&37.2\\
           sarashina2.2-3b-instruct &46.3&36.7&45.8&\textbf{56.2}&\textbf{56.6}&37.8&41.5&29.2&48.6&46.2\\
           Qwen2.5-7B-Instruct  &\textbf{69.1}&18.2&52.9&54.3&\textbf{65.0}&46.6&47.4&49.4&55.7& 53.9\\
           Meditron3-Qwen2.5-7B  &\textbf{69.1}&24.0&54.4&57.5&\textbf{63.8}&47.4&49.1&45.1&54.0&54.1\\
           Llama-3-Swallow-8B-Instruct-v1&46.0&26.4&45.6&\textbf{56.1}&\textbf{47.3}&31.8&34.6&30.2&46.5&42.6\\
           Llama-3.1-Swallow-8B-Instruct-v3&56.4&18.8&48.5&\textbf{57.5}&\textbf{56.9}&42.1&39.4&34.6&49.7&48.2\\
          Llama-3.1-Swallow-70B-Instruct-v1& \textbf{{81.}7} & {41.4} & 71.2 & 70.0 & \textbf{{82.}1} & 71.1 & 66.5 & 55.5 & 68.6& 70.9 \\ 
          Qwen2.5-72B-Instruct & \textbf{89.8}&	{51.5}&	72.2&	72.5&	\textbf{84.4}	&76.4&	68.7&	62.8&	70.0&73.6\\  \midrule
          GPT-4o & \textbf{94.4} & 76.1 & 80.9 & 83.4 & \textbf{92.1} & 88.7 & 81.8 & {72.6} & 78.6& 83.6\\ 
          o1-preview& \textbf{93.3} & 88.3 & 88.1 & {83.3} & \textbf{93.2} & 90.8 & 85.0 & 89.1 & 84.5& 87.9 \\ \midrule
            \textsc{JPharmatron}-7B /2B tokens &\textbf{80.9}&28.4&55.9&66.6&\textbf{71.5}&55.7&55.1&55.2&58.6&60.7\\ 
            \textsc{JPharmatron}-7B /10B tokens &\textbf{70.8}	&19.3	&53.6	&57.3	&\textbf{66.9}	&46.2	&48.8	&51.7	&55.3	&54.8\\
            \textsc{JPharmatron}-7B /9B tokens &\textbf{80.5}&	45.7&	57.9&	63.8&	\textbf{73.8}&	58.4	&54.9&	51.6&	61.3&	62.0\\   \bottomrule
    \end{tabular}
    }
    \caption{\textbf{Accuracy of YakugakuQA comparison by category.} Each value shows the accuracy (\%). The top two categories for each model are highlighted in bold. Most models excel in biology and pathology.}
    \label{tab:yakugakuqa-by-category}
\end{table*}

\subsection{Error analysis} \label{error-analysis}
We analyze the 16.4\% of incorrectly answered questions on YakugakuQA to identify common failure patterns and inform future improvements in domain-specific LLMs such as \textsc{JPharmatron}.

\begin{figure*}[t]
  \centering
  \begin{subfigure}[b]{0.3\textwidth}
    \includegraphics[width=\linewidth]{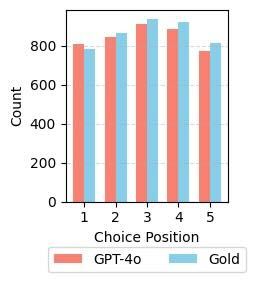}
    \caption{Positional bias (count)}
    \label{fig:position1}
  \end{subfigure}
  \hfill
  \begin{subfigure}[b]{0.33\textwidth}
    \includegraphics[width=\linewidth]{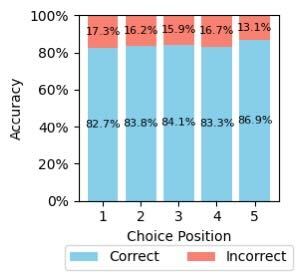}
    \caption{Positional bias (error rate)}
    \label{fig:position2}
  \end{subfigure}
  \hfill
  \begin{subfigure}[b]{0.33\textwidth}
    \includegraphics[width=\linewidth]{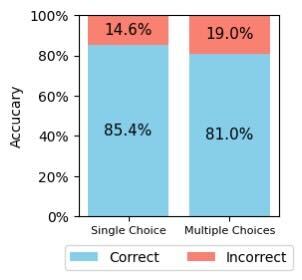}
    \caption{Single-choice vs. Multiple-choice}
    \label{fig:choices}
  \end{subfigure}\\
  \begin{subfigure}[b]{0.66\textwidth}
    \includegraphics[width=\linewidth]{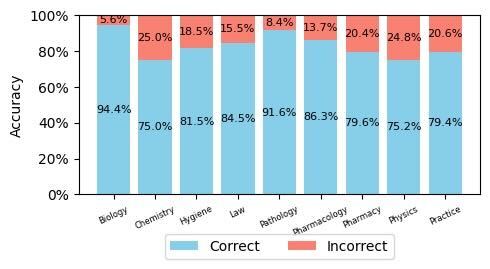}
    \caption{Category-wise accuracy}
    \label{fig:genre}
  \end{subfigure}
  \hfill
  \begin{subfigure}[b]{0.33\textwidth}
    \includegraphics[width=\linewidth]{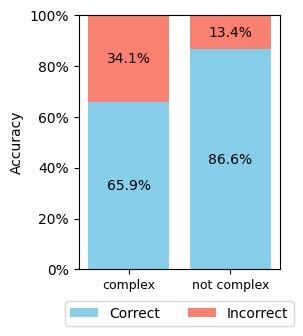}
    \caption{Complex questions}
    \label{fig:complex}
  \end{subfigure}
  \caption{\textbf{Error analysis on GPT-4o's responses in YakugakuQA.}}
\end{figure*}

\paragraph{Positional Bias.}
Consistent with previous works~\cite{marchisio-etal-2024-quantization, trung-etal-2024-reft}, we observed a positional bias in GPT-4o's responses on YakugakuQA, where the model exhibited a tendency to favor the first answer choice. Specifically, the number of responses selecting option ``1'' exceeded the total number of questions (Figure~\ref{fig:position1}), and the error rate for option ``1'' was the lowest among all choices (Figure~\ref{fig:position2}).
\paragraph{Single vs. Multiple-Choice Question.}
GPT-4o exhibited a 4.4\% higher error rate on multiple-choice questions compared to single-answer questions (Figure~\ref{fig:choices}).
\paragraph{Question category.}
Figure~\ref{fig:genre} shows that error rates for chemistry and physics are around 25\%, while those for biology and pathology are below 10\%. This indicates that GPT-4o performs better in biology and pathology, but struggles with calculation-heavy questions in chemistry and physics~\cite{ahn-etal-2024-large, li-etal-2024-gsm}. The higher performance in biology and pathology may be attributed to the prevalence of fact-based, single-answer questions in these domains. This pattern is commonly observed across various LLMs, as shown in Table~\ref{tab:yakugakuqa-by-category}, and also in JMMLU as shown in Table~\ref{tab:jmmlu-by-category}.

\paragraph{Complex questions.}
Based on the previous observation, we employed Qwen2.5-72B-Instruct~\cite{yang2024qwen2} to annotate questions requiring complex reasoning or calculations, following the LLM-as-a-Judge framework~\cite{li2024llms}. Although such questions accounted for fewer than 500 out of approximately 3000, they exhibited an error rate of 34.1\% (Figure~\ref{fig:complex}). These results suggest that top-tier LLMs still struggle with calculation-intensive tasks within the pharmaceutical domain.

\section{Discussion}

\subsection{Impact of our Benchmark Suite}

Our benchmark suite is designed to evaluate a diverse range of language capabilities required for pharmaceutical NLP. While prior datasets such as IgakuQA and JMMLU primarily focus on factual recall, our benchmarks target additional competencies that better reflect the demands of real-world pharmaceutical decision-making.

Evaluation results confirm that this broader scope offers meaningful insights. YakugakuQA and NayoseQA showed consistent improvements across most models, suggesting that domain-specific pretraining effectively enhances factual recall and term-level understanding. In contrast, SogoCheck presented a more difficult challenge. Some models showed minor gains, while others failed to improve. As previously shown, the suprisingly low accuracy of GPT-4o indicates that current LLMs --- even the state-of-the-art --- struggle with subtle consistency checks in Japanese pharmaceutical contexts.

These findings highlight the diagnostic value of SogoCheck. Rather than being a standard QA task, it probes semantic understanding capabilities that go beyond surface-level knowledge. This suggests that inconsistency detection, especially in high-stakes domains like pharmacovigilance, requires capabilities not well-captured by general LLMs.

\subsection{Deployable Domain-Specific Models: Challenges and Prospects} \label{deployable}

This study demonstrates the feasibility of building a high-performing, domain-specific LLM in Japanese without relying on commercial APIs. In pharmaceutical settings, where both data sensitivity and operational cost are critical concerns, locally trainable models such as ours present a practical and privacy-conscious alternative. Our open-source setup offers a replicable framework for enterprises and research groups seeking to train or fine-tune specialized models within secure environments.
Moreover, our benchmark suite lays the groundwork for more practical evaluations of language models in healthcare and pharmaceutical contexts. In particular, tasks like SogoCheck capture practical detection abilities that are not assessed by conventional QA benchmarks, thereby suggesting promising directions for future model and dataset development.

Despite these advances, the deployment of domain-specific models faces a critical scalability-performance tradeoff. On one hand, 7B-parameter models such as JPharmatron are relatively feasible to deploy using a small cluster of GPUs. On the other hand, such models inevitably fall short of the performance levels achieved by larger models (e.g., 70B). Bridging this gap without compromising deployability remains an open challenge, and we believe our work represents a meaningful first step toward addressing this dilemma.

Our ultimate goal in this field is to achieve a strong and useful pharmaceutical LLM. To this end, we need to further strengthen open models, as commercial models are often unavailable or restricted by regulations. 
Our experimental results, particularly those discussed in \S\ref{error-analysis}, suggest three directions for future work, listed in order of priority: (i) improving performance in core subjects to reach parity with commercial models, (ii) enhancing the overall capabilities of LLMs, and (iii) addressing weaknesses in lower-performing subjects.
While the best open models already achieve acceptable performance, they still lag clearly behind commercial counterparts (Table~\ref{tab:yakugakuqa-by-category}). As a next step, it is essential to evaluate how much performance can be improved in targeted subject areas, depending on the intended application of the model, by simply incorporating a substantial amount of relevant training data.
For the lower-performing subjects, including the improvement in chemistry and physics, both domain knowledge and reasoning ability must be significantly strengthened. However, considering development costs, we argue that addressing these weaknesses may not be a high priority in practice, as they can often be circumvented by limiting the task scope from application sides.

\section{Conclusion}

We presented \textbf{JPharmatron}, a Japanese domain-specific LLM for the pharmaceutical field, trained via continual pretraining on a bilingual pharmaceutical corpus. Alongside the model, we introduced \textbf{JPharmaBench}, the first benchmark suite covering diverse pharmaceutical language tasks.
Our model outperforms existing open medical LLMs across diverse pharmaceutical tasks, highlighting that general medical specialization alone is insufficient for pharmaceutical applications. Notably, the benchmark includes tasks such as SogoCheck, which reflect real-world document validation workflows unique to the pharmaceutical domain.
Beyond releasing a domain-specific model and benchmark, our work demonstrates the feasibility of building cost-effective, specialized LLMs deployable in secure, resource-constrained environments, which is critical for real-world use in privacy-sensitive domains like pharmaceuticals.

\section{Limitations}

\subsection*{Lack of Complete Instruction-Following Ability in LLMs}

Some smaller models tend to deviate from instructions, often generating output that includes extraneous text beyond the expected format. A common error is the inclusion of additional phrases or explanations following a colon or line break. To ensure a fair comparison in our experiments, we post-processed model outputs by extracting only the selected choice and discarding any extra text.

\subsection*{Limitations of YakugakuQA} 

Firstly, questions with images should be addressed. In particular, the chemistry category lacks sufficient coverage due to the high proportion of image-based questions. While the rise of multimodal models, especially vision-language models, is an important development, this study focuses exclusively on text-only large language models. Therefore, image-based questions were excluded from our evaluation. In the future, this limitation should be revisited when assessing multimodal models.

Moreover, YakugakuQA is a simple five-choice question-answering task, which may not be sufficient for practical implementation, although it could serve as a minimum requirement. 

Last but not least, the prompting strategy can also be improved. In our work, we used a simple setup as an initial step in this field.
It should be noted that in-context learning of LLMs has the potential to boost performance, as demonstrated by Medprompt~\cite{nori2023can} in medical question-answering for example. This point remains controversial~\cite{nori2024medprompt} and was not addressed in this study.

\subsection*{Limitations of NayoseQA}

Although we introduce a novel benchmark NayoseQA, its current format is limited to multiple-choice QA. While this format enables controlled evaluation, it may not fully reflect the practical needs of real-world entity normalization systems, where open-ended or instruction-following formats are more appropriate. To address this, we have separately released an instruction-style (SQuAD~\cite{rajpurkar2016squad}-type) variant of NayoseQA, which is not included in the main results but may serve as a valuable resource for future work on more realistic applications.

\subsection*{Limitations of SogoCheck}

SogoCheck is currently limited in scale, with only a small number of consistency pairs included in the benchmark. This restricts the statistical robustness of evaluation and may limit its confidence across different model types and domains.
In addition, generating realistic inconsistencies is inherently challenging. While we employed LLM-based generation methods to create contradictory statement pairs, it remains difficult to simulate subtle, human-like inconsistencies that naturally occur in real-world pharmaceutical texts. Many automatically generated inconsistencies tend to be either too trivial or too artificial, reducing their diagnostic value. Developing more authentic and diverse inconsistency examples remains an open challenge for future work.

\section{Acknowledgement}
This paper is based on results obtained from GENIAC (Generative AI Accelerator Challenge, a project to strengthen Japan’s generative AI development capabilities), a project implemented by the Ministry of Economy, Trade and Industry (METI) and the New Energy and Industrial Technology Development Organization (NEDO).

We thank Yuki Kobiyama, Kouta Hiroe, Masabumi Ishihara, and Miyuki Toyoi for helpful supports. 
We used ChatGPT-4o for the support of proofreading.

\bibliography{custom}

\appendix

\section{Ethical considerations}
While \textsc{JPharmatron} is designed to complete pharmaceutical tasks resembling the real tasks in pharmacy companies, it is not yet confirmed to accomplish the real tasks within professional acceptable quality. 
It raises several ethical considerations that must be addressed to ensure responsible development and deployment.

Importantly, the model may still generate factually incorrect or misleading content. 
We recommend to further finetune our model with the company's real data and conduct additional use-case alignment and testing before deploying it in real-world practice.
We further emphasize that the model is not intended for clinical use. Instead, it is suitable for document processing tasks, where potential risks can be mitigated through human review and validation of the generated content.

The training data may contain biases related to demographics, geographic representation, or commercial interests. Additionally, if any data were to originate from patents, proprietary databases, or unpublished sources, there would be a risk of inadvertently disclosing protected content or facilitating unauthorized reuse. Although all training data used in this study were sourced from publicly available datasets, we acknowledge that this issue was not directly addressed in the current work.

\section{Supplementary information on our benchmarks} \label{appendix:benchmarks}
\subsection{YakugakuQA}
The number of YakugakuQA is listed in Table~\ref{tab:number-of-yakugakuqa}. Among the available questions online, only those with texts were extraceted.

\begin{table*}[!t]
    \centering
    \resizebox{\linewidth}{!}{
    \begin{tabular}{cccccccccc|c}
    \toprule
        & \textbf{Biology} & \textbf{Chemistry} & \textbf{Hygiene} & \textbf{Law} & \textbf{Pathology} & \textbf{Pharmacology} & \textbf{Pharmacy} & \textbf{Physics} & \textbf{Practice} & \textbf{Total} \\ \midrule
        2012 & 17 & 4 & 30 & 29 & 37 & 38 & 36 & 17 & 65 & 273 \\ 
        2013 & 16 & 3 & 32 & 28 & 36 & 34 & 33 & 11 & 63 & 256 \\ 
        2014 & 15 & 4 & 28 & 29 & 35 & 37 & 28 & 13 & 63 & 252 \\ 
        2015 & 8 & 3 & 26 & 27 & 35 & 35 & 31 & 9 & 60 & 234 \\ 
        2016 & 10 & 3 & 30 & 27 & 37 & 40 & 29 & 12 & 50 & 238 \\ 
        2017 & 11 & 2 & 28 & 26 & 37 & 36 & 27 & 10 & 54 & 231 \\ 
        2018 & 11 & 4 & 31 & 27 & 36 & 35 & 25 & 10 & 53 & 232 \\ 
        2019 & 9 & 1 & 28 & 28 & 32 & 33 & 26 & 12 & 46 & 215 \\ 
        2020 & 12 & 4 & 25 & 26 & 33 & 33 & 17 & 12 & 42 & 204 \\ 
        2021 & 6 & 2 & 30 & 27 & 35 & 30 & 19 & 10 & 55 & 214 \\ 
        2022 & 9 & 3 & 25 & 27 & 33 & 33 & 24 & 15 & 48 & 217 \\ 
        2023 & 10 & 3 & 23 & 25 & 27 & 33 & 22 & 15 & 47 & 205 \\ 
        2024 & 11 & 11 & 33 & 23 & 28 & 36 & 31 & 18 & 59 & 250 \\ \bottomrule
    \end{tabular}
    }
    \caption{\textbf{The number of questions used in our experiments by year and category.} The questions that include images have been excluded from the original NPLE.}
    \label{tab:number-of-yakugakuqa}
\end{table*}

\paragraph{Prompt}
Below are the three-shot examples included in the prompt throughout our experiments. All of them are originally in Japanese, but translated into English by ChatGPT-4o mini for this article.

\begin{quote}
Question: Which of the following insomnia medications inhibits the orexin receptor?
Please select exactly one from the options 1, 2, 3, 4, or 5.\\
1: Brotizolam\\
2: Flunitrazepam\\
3: Eszopiclone\\
4: Ramelteon\\
5: Lemborexant\\
Answer: 5\\
Question: Which two mechanisms of action describe the effects of sacubitril/valsartan?
Please select exactly two from the options 1, 2, 3, 4, or 5.\\
1: Inhibits neprilysin, thereby preventing the breakdown of endogenous natriuretic peptides, resulting in vasodilation and diuretic effects.\\
2: Inhibits angiotensin II receptors, suppressing aldosterone secretion from the adrenal cortex, thereby causing vasodilation.\\
3: Acts on ANP receptors in the blood vessels and kidneys, activating guanylate cyclase, resulting in vasodilation and diuretic effects.\\
4: Blocks aldosterone receptors in the collecting ducts, leading to diuretic effects.\\
5: Inhibits angiotensin-converting enzyme, thereby preventing the formation of angiotensin II, resulting in vasodilation.\\
Answer: 1,2\\
Question: Which of the following migraine prophylactic drugs inhibits calcitonin gene-related peptide (CGRP)?\\
Please select exactly one from the options 1, 2, 3, 4, or 5.\\
1: Basiliximab\\
2: Trastuzumab\\
3: Benralizumab\\
4: Galcanezumab\\
5: Tocilizumab\\
Answer: 4\\
\end{quote}

\subsection{Pharmaceutical-related subset of JMMLU} \label{appendix:jmmlu}

The number of questions included in each category of JMMLU which was used in our evaluation experiments is listed in Table~\ref{tab:jmmlu-num}. The category-wise accuracy is shown in Table~\ref{tab:jmmlu-by-category}. Consistent with the results in YakugakuQA (Table~\ref{tab:yakugakuqa-by-category}), the overall trend that biology tends to score higher than chemistry and physics is observed.

\begin{table}[]
    \centering
    \resizebox{0.9\linewidth}{!}{
    \begin{tabular}{cc}
    \toprule
        \textbf{Category} & \textbf{The number of questions}\\ \midrule
         clinical\_knowledge&150	\\
         college\_biology&143\\
         college\_chemistry&	99\\
         college\_medicine&	150	\\
         college\_physics&100\\
         high\_school\_biology&	148	\\
         high\_school\_chemistry&149	\\
         high\_school\_physics&150	\\
         high\_school\_statistics&150\\
         medical\_genetics&	99	\\
         nutrition&149\\
         professional\_medicine&	150\\
         virology &150\\\midrule
         Total& 1787\\ \bottomrule
    \end{tabular}}
    \caption{\textbf{The number of questions by categories included in pharmaceutical-related JMMLU.}}
    \label{tab:jmmlu-num}
\end{table}

\begin{table*}[!ht]
    \centering
    \resizebox{\linewidth}{!}{
    \begin{tabular}{lccccccccccccc|c}\toprule
          \textbf{Model}& \textbf{clinical\_}& \textbf{college\_}& \textbf{college\_}& \textbf{college\_}& \textbf{college\_}& \textbf{high\_school\_}& \textbf{high\_school\_}& \textbf{high\_school\_}& \textbf{high\_school\_}& \textbf{medical\_}& \textbf{nutrition}& \textbf{professional\_}& \textbf{virology}& \textbf{Over}\\ 
          &\textbf{knowledge}&\textbf{biology}&\textbf{chemistry}&\textbf{medicine}&\textbf{physics}& \textbf{biology}&\textbf{chemistry}&\textbf{physics}&\textbf{statistics}&\textbf{genetics}&&\textbf{medicine}&&\textbf{-all}\\ \midrule
          TinySwallow-1.5B-Instruct &41.3&28.0&29.3&36.0&28.0&40.5&26.8&25.3&28.7&31.3&34.2&30.7&34.0&32.1\\
           sarashina2.2-3b-instruct &39.3&45.5&29.3&42.0&35.0&52.7&26.2&27.3&34.0&40.4&47.7&44.7&24.7&37.8\\
           Qwen2.5-7B-Instruct  &52.7&46.9&30.3&41.3&37.0&50.7&36.2&28.7&32.7&48.5&57.7&49.3&41.3&42.9\\
           Meditron3-Qwen2.5-7B  &48.7&27.3&19.2&26.7&33.0&37.8&23.5&28.7&34.7&28.3&44.3&33.3&22.0&31.7\\
           Llama-3-Swallow-8B-Instruct-v0.1&30.7&12.6&17.2&25.3&11.0&26.4&20.1&21.3&27.3&11.1&16.1&30.0&11.3&20.6\\
           Llama-3.1-Swallow-8B-Instruct-v0.3&52.0&45.5&35.4&47.3&37.0&55.4&35.6&30.0&36.7&55.6&53.7&44.7&42.0&44.0\\ \midrule
           GPT-4o&82.7&	93.0&	60.6&	81.3&	69.0&	85.1	&76.5&	70.0	&82.0	&88.9	&82.6	&94.7	&56.7&	79.1\\ \midrule
           Ours (best)& 58.7	&64.3	&44.4	&48.7	&50.0&	65.5	&48.3	&46.0	&64.7	&59.6	&62.4	&58.7	&40.7	&55.0\\ \bottomrule
    \end{tabular}
    }
    \caption{\textbf{Accuracy comparison on JMMLU across different subject categories and different LLMs.} }
    \label{tab:jmmlu-by-category}
\end{table*}

\section{Model \& Training} \label{appendix:model_training}

\subsection{Data accumulation}
The continual pretraining corpus used for \textsc{JPharmatron} is composed of five categories of text, collected from publicly available sources. Each data type was selected to contribute domain-relevant knowledge or general linguistic fluency. An overview is provided below:

\paragraph{Journal Articles}  
Academic papers and review articles related to pharmacology, pharmacy practice, and clinical medicine. These texts provide rich domain-specific vocabulary and formal written structures.

\paragraph{PubMed Abstract Subset}  
A curated selection of English abstracts from the PubMed database, focusing on drug-related publications. This source contributes approximately 8 billion tokens and provides a biomedical foundation to complement the Japanese data.

\paragraph{Package Inserts approved by PMDA}  
Texts published by Japan's Pharmaceuticals and Medical Devices Agency (PMDA), such as drug approval summaries, review reports, and safety alerts. These documents contribute approximately 87 million tokens and reflect regulatory terminology.

\paragraph{Official Documents from Governmental Institutes}  
Documents from government-affiliated organizations including the Pharmaceuticals and Medical Devices Act.

\paragraph{General-Domain Corpus}  
A part of FineWeb\footnote{\url{https://huggingface.co/datasets/HuggingFaceFW/fineweb}} and Swallow Dataset\footnote{\url{https://huggingface.co/datasets/tokyotech-llm/swallow-magpie-ultra-v0.1}}.

\subsection{Data Filtering}
We constructed a high-quality, domain-specific corpus for the pharmaceutical domain by leveraging a multi-stage filtering pipeline built upon large language models (LLMs) and trained classifiers. Following SmolLM2~\cite{allal2025smollm2}, the overall procedure consists of three steps:
\begin{enumerate}
    \item We first sampled a subset of documents from the Common Crawl dataset (CC100). A high-performing LLM (Qwen2.5-72B) was prompted to assign each page a pharmaceutical relevance score ranging from 0 (irrelevant) to 5 (highly relevant).
    \item Using 54,056 LLM-labeled samples, we trained a classifier to predict the pharmaceutical relevance score of input documents. Pages scoring 1 or higher were retained.
    \item The retained documents were further evaluated using the same LLM to assign an educational quality score (0-5). A second classifier, trained on 5,478 LLM-labeled samples, was used to filter out documents with an educational quality score 3 or lower. This ensured that the resulting data not only pertains to pharmaceutical content but is also of pedagogical value.
\end{enumerate}
All training data for both classifiers were generated using high-confidence outputs from the Qwen2.5-72B model. Both classifiers were trained following the configuration of the finemath-classifier\footnote{\url{https://huggingface.co/HuggingFaceTB/finemath-classifier}} framework.

As a result of this filtering pipeline, we collected 904,651 high-quality, pharmaceutical-related documents (totalling 1.2 billion tokens) from the deduplicated Common Crawl (llm-jp-corpus-v3\footnote{\url{https://gitlab.llm-jp.nii.ac.jp/datasets/llm-jp-corpus-v3}}).

\subsection{Data cleansing}

In this study, we employed the D4 algorithm~\cite{tirumala2023d4} to perform data deduplication, aiming to reduce redundant information. D4 is primarily composed of SemDeDup (Semantic deduplication)~\cite{abbas2023semdedup} and 
SSL Prototype (Self-Supervised Learning Prototypes)~\cite{sorscher2022beyond}. The former incorporates $k$-means clustering to eliminate texts with cosine similarity larger than $1-\epsilon$. 
We set $\epsilon = 3\times 10^{-8}$ for the discarding threshold in SemDeDup and $R=0.95$ for the discarding proportion in SSL Prototype, respectively. 
In summarization, the total number of tokens were reduced from 10B to 9B.

\subsection{Base model selection}

Discussing industrial applications often lead to the cost perspectives. Different from research purpose development, the operational cost in inference phase also should be taken into account, otherwise no institution can afford to utilize the trained model. 
Training a model from scratch to learn Japanese was deemed prohibitively costly. 
Therefore, in selecting the base model, we prioritized the use of a pretrained model that had already been trained on Japanese data, and we also sought a model with a commercially viable license that would facilitate its adoption within the pharmaceutical industry. 
We restricted the model size to around 7B for better usability considering the training cost and inference cost. 
Based on these criteria, we chose Qwen2.5-7B~\cite{yang2024qwen2} as the base model.

\subsection{Enhancing Instruction Following via Model Merging} \label{appendix:model merging}

Our domain-specific model trained through continued pretraining exhibited poor instruction-following capabilities. As a result, these models struggle to answer multiple-choice questions correctly, rendering them ineffective for standard benchmark evaluations which rely heavily on such tasks.

Instead of applying supervised fine-tuning (SFT), which can be resource-intensive and require carefully aligned datasets, we adopt a lightweight approach by leveraging model merging. Specifically, we aim to endow a domain-adapted model with strong instruction-following and reasoning capabilities by merging it with a general-purpose instruction-tuned model.

To this end, we designate Qwen2.5-7B-Instruct as the base model, given its demonstrated strength in instruction adherence and task generalization. The domain-specific model, pretrained on 2B tokens of pharmaceutical texts, serves as the knowledge-rich counterpart in the merge.

We employ the TIES merging strategy~\cite{yadav2023ties} provided by \textit{mergekit}~\cite{goddard-etal-2024-arcees}, and assign a weight to balance the retention of domain knowledge while preserving the core reasoning and output structure of the instruction-tuned base model. Table~\ref{tab:model merging} shows the superiority of EvoLLM~\cite{akiba2025evolutionary} coupled with DARE TIES merging.

\begin{table}
    \centering
    \resizebox{0.9\linewidth}{!}{
    \begin{tabular}{cc} \toprule
         Merge method& YakugakuQA (\%)\\ \midrule
         TIES (weight 8:2)& 57.2\\ 
         TIES (weight 7:3)& 59.0\\ 
         TIES (weight 6:4)& 60.4\\ 
         DARE TIES by EvoLLM& \bf{60.7}\\ \bottomrule
    \end{tabular}
    }
    \caption{\textbf{Accuracy comparison on YakugakuQA across different merging methods.} Qwen2.5-7B-Instruct was used as the base model and \textsc{JPharmatron}-7B (Ours) was used as the auxiliary model.}
    \label{tab:model merging}
\end{table}

\end{document}